\documentclass{article}

\usepackage{PRIMEarxiv}

\usepackage[utf8]{inputenc}
\usepackage[T1]{fontenc}
\usepackage{amsmath,amsfonts,amssymb}
\usepackage{algorithm}
\usepackage{algorithmic}
\usepackage{array}
\usepackage{booktabs}
\usepackage{caption}
\usepackage{cite}
\usepackage{colortbl}
\usepackage{float}
\usepackage{graphicx}
\usepackage{makecell}
\usepackage{microtype}
\usepackage{multirow}
\usepackage{nicefrac}
\usepackage{textcomp}
\usepackage{xcolor}
\usepackage{url}
\usepackage{hyperref}
\hypersetup{hidelinks}

\graphicspath{{figures/}}

\title{CarbonCLIP: Enhance Carbon Prediction from Satellite Imagery via Integrated Street-View Semantics and Temporal Context Training}

\author{
Zeru Yang$^{1,2}$, Fang-Ying Gong$^3$, Steve H.L. Yim$^{4,5}$, Chau Yuen$^{2,5}$\\
$^1$Energy Research Institute at NTU, Interdisciplinary Graduate Programme, Nanyang Technological University, Singapore\\
$^2$School of Electrical and Electronic Engineering, Nanyang Technological University, Singapore\\
$^3$School of Public Administration and Policy, Renmin University of China, Beijing, China\\
$^4$Asian School of the Environment, Nanyang Technological University, Singapore\\
$^5$Center for Climate Change and Environmental Health, Nanyang Technological University, Singapore\\
\texttt{zeru001@e.ntu.edu.sg, fangying.gong@ruc.edu.cn, steve.yim@ntu.edu.sg, chau.yuen@ntu.edu.sg}
}

\begin{document}
\maketitle

\begin{abstract}
Accurately estimating urban carbon emissions is critical for sustainable urban planning, yet many existing approaches remain difficult to apply consistently across cities due to data-source heterogeneity and the lack of fine-grained semantic-temporal context in remote sensing data. 
We propose CarbonCLIP, a task-oriented multimodal distillation framework that improves satellite-based carbon emission prediction by transferring contextual knowledge into a unified satellite representation through dual-branch contrastive learning. 
Unlike conventional methods that rely on static visual features, CarbonCLIP explicitly bridges the gap between top-down satellite views and ground-level human activities. 
Specifically, the spatial branch uses fine-grained textual descriptions automatically generated from street-view images by Large Multimodal Models (LMMs) to provide semantic priors reflecting building functions, infrastructure, and urban activities, while the temporal branch employs a month encoder to encode temporal priors associated with monthly emission variation. 
CarbonCLIP requires multimodal data only during the pretraining phase; during inference, it relies solely on satellite imagery, thereby supporting scalable deployment when ground-level data are unavailable at inference. 
Experiments on Beijing and Singapore demonstrate that CarbonCLIP outperforms baselines in both study cities. 
The results validate that our method effectively transfers multimodal knowledge into satellite representations, offering a robust solution for satellite-based urban carbon modeling.
\end{abstract}

\section{Introduction}

With the rapid urbanization occurring worldwide, cities have become the dominant sources of anthropogenic carbon emissions, accounting for more than one-third of global totals~\cite{Shen_2022, Crippa_2021}. Accurately quantifying and estimating urban carbon emissions is essential for developing sustainable and climate-resilient cities~\cite{Ahad_2020}. Urban sustainability requires not only reducing carbon footprints but also integrating intelligent monitoring tools that can inform policy-making and promote equitable environmental outcomes. A core challenge in achieving this goal lies in how we perceive and analyze cities at scale. Carbon emissions in urban areas are deeply intertwined with spatial patterns such as land use and infrastructure, as well as human-scale factors like greenery, density, and activity levels.

\begin{figure}[H]
    \centering
    \includegraphics[width=\columnwidth]{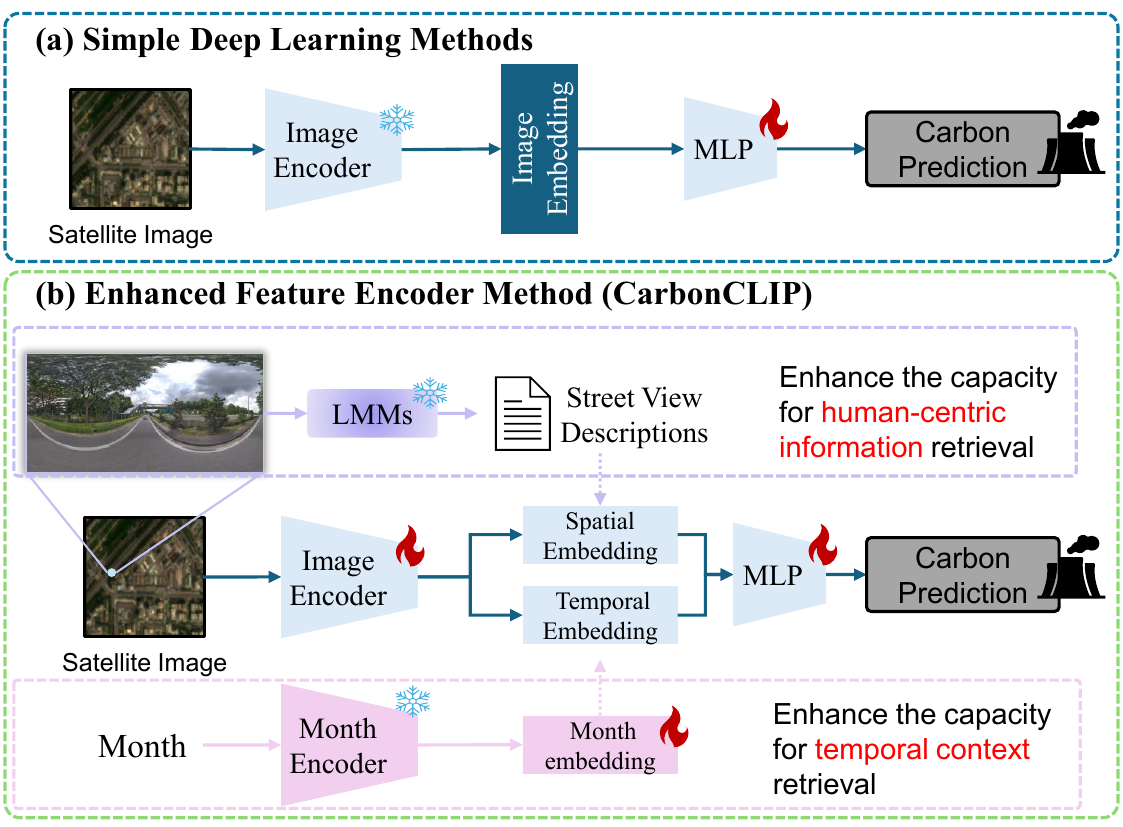}
    \caption{Comparison of frameworks for carbon emission prediction: (a) Simple transfer learning framework based on visual embeddings; (b) The proposed CarbonCLIP method enhanced via LMM-generated human-centric information from street-view imagery and temporal context with month indication.}
    \label{fig:methods_comparison}
\end{figure}

In recent years, visual data have become a cornerstone of urban analytics and environmental monitoring, providing unprecedented means of observing cities at multiple scales. Satellite imagery offers a comprehensive overview of urban environments, capturing spatial structure and physical layout~\cite{Albert_2017}. As a scalable and consistent data source, it has been widely applied in air pollution mapping~\cite{Wald_2020}, agricultural monitoring~\cite{Nakalembe_2021}, and carbon stock estimation~\cite{Reiersen_2022}. Advances in remote sensing, including high-resolution Planet imagery at 3~m resolution~\cite{planet_api}, enable detailed detection of urban expansion and environmental change~\cite{Rolf_2024}, shown in Fig.~\ref{fig:methods_comparison}(a). Yet, despite its broad coverage, satellite imagery lacks human-centric information such as facade features and street-level greenery, limiting its ability to capture fine-grained drivers of emissions and livability.

In contrast, street-level imagery from platforms like Google Maps and Baidu Maps provides a complementary ground-level perspective~\cite{Shen_2017}, revealing visual details often missed from above, including building facades, signage, sidewalks, and vegetation~\cite{Gong_2018, Fan_2023, Biljecki_2021}. These human-scale semantics relate closely to livability, mobility, and micro-environmental quality, which influence emissions at local scales~\cite{Li_2018, Pang_2022, Lu_2024}. However, their spatial and temporal sparsity restricts large-scale and continuous monitoring. The advancement of Large Multimodal Models (LMMs) provides a good foundation for extracting semantic information from visual data, and by utilizing LMMs to abstract visual street scenes into structured textual descriptions, we bypass the need for precise pixel-level geometric alignment, enabling robust knowledge transfer from the human-scale perspective to the satellite domain.

Furthermore, urban carbon emissions exhibit temporal volatility associated with seasonal energy consumption cycles (e.g., heating and cooling) and phenological changes~\cite{Yang_2024, Couture_2024, Hamilton_2024}. While satellite imagery provides periodic revisits (e.g., monthly observations) that preserve month-level temporal variation, street-view imagery is typically collected at sparse, fixed time points and represents a static moment of the urban fabric. Consequently, directly aligning satellite imagery with static street-view data may suppress temporal information in the satellite observations. This limitation motivates a dedicated mechanism for incorporating temporal context alongside spatial semantics.

Recent studies have pursued multi-source data fusion, combining points of interest (POIs), socio-demographic statistics, transportation facilities, and remote sensing data~\cite{Zheng_2022, Wu_2023, Wang_2025}. Others have begun coupling satellite and street-view imagery to enrich urban representation for downstream tasks such as carbon emission prediction~\cite{Zhang_2025, Hao_2025, Chen_2022}. Despite these advances, data-source heterogeneity, arising from inconsistent spatial, temporal, and measurement standards, hampers harmonization, and reliance on region-specific auxiliary data limits scalability. Many cities, especially in developing regions, lack reliable or frequently updated datasets, making model deployment outside the original data-rich setting uncertain.

To reduce dependence on region-specific auxiliary sources in urban carbon emission prediction, a paradigm shift is needed, from aggregating local auxiliary datasets to distilling multimodal knowledge into a satellite-centered representation. Building on this insight, we propose CarbonCLIP, a framework that enhances carbon prediction from satellite imagery by fusing street-view semantics and temporal context during training, depicted in Fig.~\ref{fig:methods_comparison}(b). Through dual-branch contrastive learning, CarbonCLIP transfers fine-grained spatial semantics and temporal priors into the satellite domain, enabling a satellite-only model to retain multimodal awareness during inference. Specifically, CarbonCLIP employs a pretrained image encoder for satellite features, uses automatically generated textual descriptions as semantic anchors, and incorporates temporal embeddings to capture monthly temporal correlations. The final predictor is trained for monthly emission regression using only satellite imagery, forming a unified and scalable modeling paradigm.

CarbonCLIP differs from previous multi-source or satellite-based methods in three key aspects. First, it bridges the gap between top-down and ground-level perspectives by aligning satellite imagery with textual descriptions automatically generated from street-view images, thereby transferring authentic human-centric semantics into a scalable satellite representation. Second, CarbonCLIP introduces a cyclic temporal alignment mechanism through month embeddings, enabling the model to encode temporal priors associated with monthly emission variation and dependencies often ignored in static visual models. Third, unlike conventional multi-source frameworks that rely on region-specific auxiliary data, CarbonCLIP distills multimodal contextual knowledge during training but performs inference using only satellite imagery, supporting deployment when auxiliary ground-level observations are unavailable at inference. In summary, our main contributions are:
\begin{itemize}
    \item We propose a contrastive learning framework that distills street-level semantics and temporal knowledge into a unified satellite representation, achieving multimodal awareness while maintaining single-modality inference.
    \item We construct a dataset by generating fine-grained textual descriptions from street-view imagery and plan to make it publicly available, facilitating future research on scalable urban understanding.
    \item We validate CarbonCLIP on two cities with distinct spatial and climatic patterns, Beijing and Singapore, showing substantial improvements in prediction accuracy in both study cities.
\end{itemize}

\section{Related Work}

\subsection{Carbon Emission Prediction}
Early studies predominantly relied on bottom-up inventory methods~\cite{Gallo_2014, Oda_2018}, which estimate emissions by aggregating energy consumption statistics with predefined emission factors. Although these approaches are widely regarded as the most reliable benchmarks, they suffer from inherently coarse spatiotemporal resolution, typically annual and city-level, and substantial reporting delays. These limitations significantly restrict their applicability for fine-grained analysis and timely urban carbon management.

To overcome these constraints, recent research has explored data-driven models that fuse multiple urban data sources, including point of interest (POI) distributions~\cite{Wei_2024, Zheng_2022}, vehicle trajectories~\cite{Cui_2023, Wu_2023, Xiao_2024}, and energy metering records~\cite{Wang_2025, Gao_2024}. By capturing human activity patterns and socioeconomic conditions, those methods achieve improved spatial granularity. However, their reliance on region-specific data severely limits scalability and makes deployment beyond the original data-rich setting uncertain. The availability, quality, and definition of fine-grained socioeconomic datasets vary substantially across regions, making these models difficult to apply consistently, especially to data-scarce or developing regions.

Remote sensing offers a promising alternative due to its global coverage, consistency, and independence from local statistical infrastructures. Initial remote-based studies employed night-time light (NTL) imagery as a proxy for economic activity to spatially disaggregate carbon emissions~\cite{Wei_2024, Wang_2024, Fang_2022}. While effective at large scales, NTL data are prone to saturation effects in dense urban cores~\cite{Hu_2021,Li_2018, Sun_2024} and lack the spatial resolution needed to distinguish detailed emission sources. With advances in deep learning, more recent work has shifted toward high-resolution daytime optical satellite imagery (e.g., Sentinel-2, Planet)~\cite{Kim_2021, Hobbs_2023}, leveraging urban morphological information (such as building density, vegetation coverage, and road networks)~\cite{Couture_2024, Ojadi_2024, Liu_2024} for carbon emission prediction.

Despite these advances, remote-based approaches face a fundamental limitation: optical satellite imagery primarily captures spectral reflectance and surface texture, which are often insufficient to distinguish functionally distinct urban zones with similar visual appearances from a top-down perspective (i.e., iso-spectral objects)~\cite{Chen_2022}. Moreover, satellite observations inherently lack vertical and ground-level information, such as building facade characteristics, business types inferred from signage, and street-level traffic intensity, which are factors that are strongly correlated with energy consumption and carbon emissions~\cite{Darabi_2025,Xiao_2024, Shen_2017, Biljecki_2021}.

These challenges motivate a satellite-centered method that can inject fine-grained ground-level semantics and temporal contextual knowledge into satellite-based representations, while keeping inference independent of region-specific auxiliary data.

\subsection{Cross-Modal Geospatial Representation Learning}
Recent geospatial representation learning increasingly uses contrastive alignment to connect visual observations with external contextual signals~\cite{Li_2023, Sun_2025}. One common strategy is to use geographic coordinates as dense and easily available supervision. GeoCLIP~\cite{Vivanco_2023} aligns natural images with GPS locations for image geo-localization, whereas SatCLIP~\cite{Klemmer_2025} contrastively matches Sentinel-2 satellite observations with coordinates to learn a global geographic location encoder. Coordinate supervision is scalable because location labels are widely available, but it mainly tells the model where an observation is located, not what urban functions or emission-related activities are present there. This distinction is important for carbon emission modeling: nearby or geographically similar areas may contain different land uses, infrastructure densities, and activity intensities, whereas the same urban function may appear in different districts~\cite{Zhou_2024, Jia_2025}. This motivates semantic supervision that describes what is present in the scene, not only where the scene is located.

To mitigate these limitations, vision--language geospatial models introduce explicit semantic supervision into remote-sensing and urban representations. RemoteCLIP~\cite{Liu_2024_Remoteclip} addresses remote-sensing pretraining-data scarcity by converting heterogeneous annotations into image-caption pairs through box-to-caption (B2C) and mask-to-box (M2B) conversions, then continually pretraining CLIP for remote-sensing vision-language representation. UrbanCLIP~\cite{Yan_2024} instead targets urban region profiling: it generates satellite-image descriptions with an image-to-text LLM and trains image-text representations with contrastive and language-modeling losses for urban indicator prediction. Earlier RS-CLIP-style models~\cite{Li_2023_RS} and remote-sensing multimodal large language models such as GeoChat~\cite{Kuckreja_2024}, EarthGPT~\cite{Zhang_2024}, and EarthGPT-X~\cite{Zhang_2025_EarthGPT-X} further show that language supervision can support diverse remote-sensing understanding tasks, including scene classification, captioning, visual question answering, visual grounding, and object detection. More recent advances further enrich semantic representations by incorporating ground-level street-view imagery~\cite{Hao_2025, Zhang_2025, Xu_2025}. By bridging overhead satellite observations with human-scale perspectives, these cross-view frameworks aim to capture fine-grained physical and functional attributes that are invisible from a top-down view alone. Nevertheless, most existing approaches still rely on retrieval-based supervision or static feature alignment, and thus fail to fully exploit generative semantics, wherein LMMs can synthesize dense and structured descriptions of urban scenes.

\begingroup
\color{black}
\begin{table}[!t]
    \centering
\caption{Descriptive comparison of representative CLIP-style geospatial representation methods. The table summarizes each method by the representation it learns and the training signal it uses rather than ranking them by a single criterion.}
    \label{tab:method_positioning}
    \setlength{\tabcolsep}{3pt}
    \begin{tabular}{p{2.3cm} p{5cm} p{8.0cm}}
        \toprule
\textbf{Method} & \textbf{Learned representation} & \textbf{Training signal / objective} \\
\midrule
SatCLIP~\cite{Klemmer_2025} & Geographic location representation & Contrastively matches Sentinel-2 satellite imagery with geographic coordinates to learn a location encoder \\
RemoteCLIP~\cite{Liu_2024_Remoteclip} & Remote-sensing image--text representation & Continual CLIP pretraining on image-caption pairs constructed from detection, segmentation, and existing image-text datasets \\
UrbanCLIP~\cite{Yan_2024} & Text-enhanced satellite representation for urban regions & Uses LLM-generated satellite-image descriptions and optimizes image-text contrastive plus language-modeling losses \\
CarbonCLIP (ours) & Satellite representation for monthly carbon emission prediction & Distills street-view-generated text and month context into satellite features through spatial-semantic and temporal contrastive alignment \\
        \bottomrule
    \end{tabular}
\end{table}
\endgroup

Table~\ref{tab:method_positioning} summarizes these representative methods descriptively according to the representation they learn and the training signal they use. Moreover, across both coordinate-based and vision--language alignment paradigms, the temporal dimension remains largely overlooked. Urban representations are typically learned from static snapshots, neglecting seasonal variations in vegetation, energy demand, and human activity that fundamentally shape urban dynamics~\cite{Yang_2024, Hamilton_2024}. This limitation is particularly restrictive for applications such as carbon emission modeling, where temporal patterns play a central and indispensable role.

\begin{figure}[!t]
    \centering
    \includegraphics[width=\textwidth]{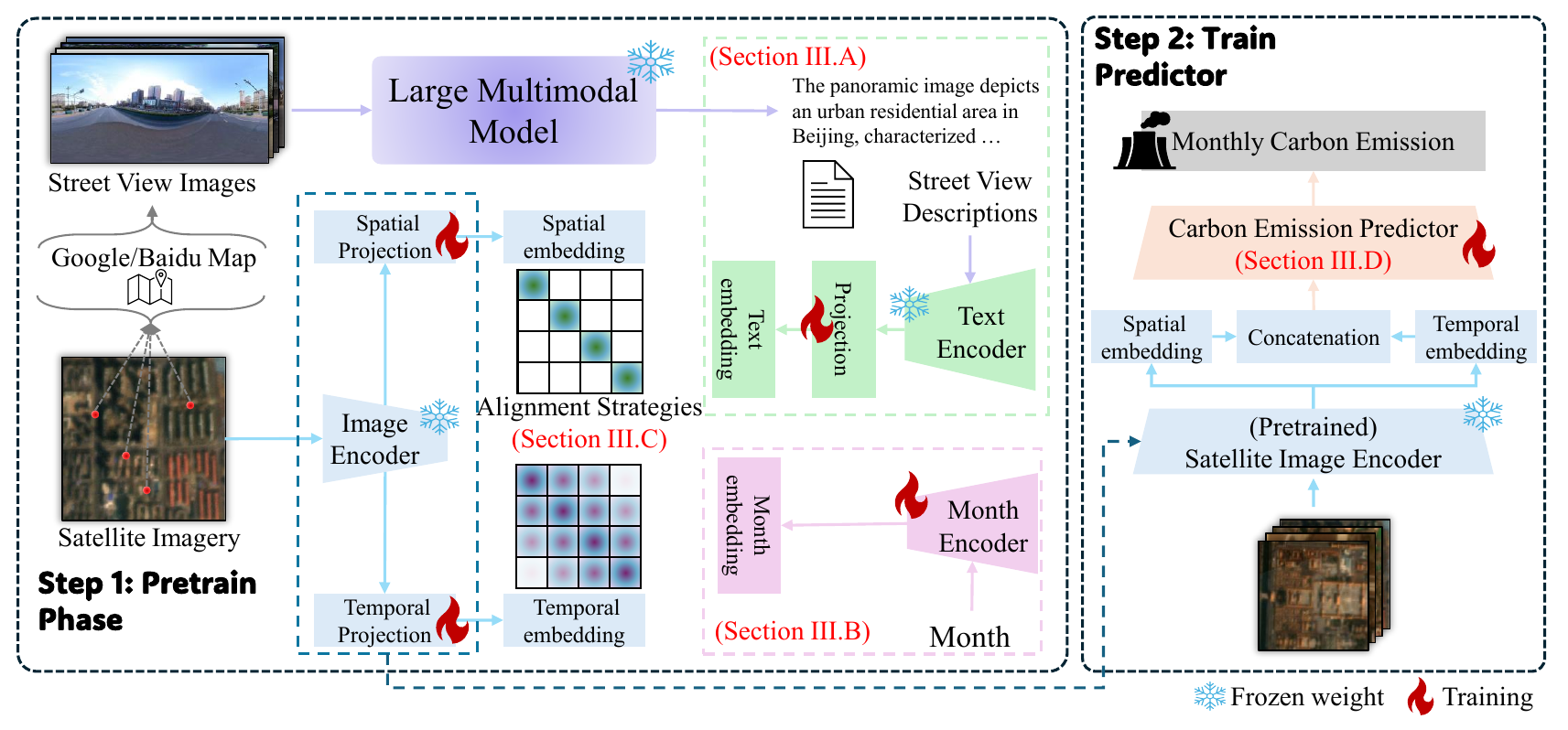} 
    \caption{Overall framework of our proposed CarbonCLIP.}
    \label{fig:framework}
\end{figure}

\section{Proposed CarbonCLIP Methodology}
The proposed CarbonCLIP architecture is shown in Fig.~\ref{fig:framework}, which enhances carbon emission prediction from satellite imagery by fusing street-view semantics and temporal context during training. An automatic pipeline first generates fine-grained textual descriptions of street-view images using both geographic and visual prompts, powered by the multimodal large language model Qwen2.5-VL \cite{Bai_2025}. In parallel, a month encoder is designed to capture temporal representations of emission patterns. During the pretraining stage, CarbonCLIP employs dual contrastive objectives to distill multimodal knowledge into the satellite representation: one objective transfers detailed spatial semantics from street-view descriptions to satellite images, while the other injects temporal variation by associating month embeddings with satellite imagery. After pretraining, the satellite image encoder and projection layers are frozen, and a lightweight multi-layer perceptron (MLP) is trained for monthly carbon emission regression using only satellite imagery, ensuring an efficient and scalable inference pipeline.

\subsection{Street-View Semantic Representation}

Providing detailed descriptions for large-scale collections of street-view images poses a significant challenge, as traditional methods such as manual annotation are labor-intensive, time-consuming, and require substantial human resources. With the rapid advancement of LMMs, however, it is now possible to automatically generate rich and context-aware textual descriptions that capture both visual and geographical semantics. In this study, we leverage Qwen2.5-VL \cite{Bai_2025}, an advanced, stable, and open-source LMM developed by Alibaba, which demonstrates capability in visual understanding tasks. To further enhance the quality and specificity of the generated street-view descriptions, we incorporate prompt strategies inspired by the UrbanCLIP model \cite{Yan_2024}. Specifically, we design structured prompts as follows that include not only visual information but also geographic metadata, such as city names and precise longitude and latitude coordinates. This integration of spatial context into the prompting process enables the model to generate more accurate, detailed, and geographically grounded descriptions.

\begin{quote}
\texttt{"You are a helpful assistant to analyze street-view images. \\
Analyze the features about urban structure and environment from the panoramic image of street-view in [city] in a comprehensive and detailed manner and summarize it into one paragraph: The coordinate of the street-view image is [longitude], [latitude]. Note that ignore the weather and other details that change over a short period of time on the image. Note that ignore the street-view vehicle and other details but focus on the overall features. Note that try to avoid inferences and focus on the content of the image."}
\end{quote}

Following the generation process, the descriptions are processed by a frozen text encoder from the CLIP framework~\cite{Radford_2021} to obtain high-dimensional semantic embeddings. Specifically, urban 1~km\textsuperscript{2} spatial tiles can contain heterogeneous land-use components, such as residential blocks, commercial streets, transport corridors, and green spaces. We therefore do not use a single street-view image as a direct tile-level label. Instead, all retained street-view panoramas whose GPS coordinates fall inside the same 1~km\textsuperscript{2} spatial tile are associated with that tile. Each panorama is first converted into a textual description and encoded by the Transformer-based text encoder. The encoder's pooled text embedding is used as the semantic representation for the corresponding panorama, and the tile-level street-view semantic representation is then constructed at the embedding level from all panoramas matched to that tile. These text embeddings serve as ground-level semantic anchors, enabling the subsequent alignment with satellite image features within a shared contrastive latent space.

\subsection{Month-Level Temporal Context Encoding}
\label{sec:Month Encoder}

To incorporate temporal information into CarbonCLIP without attributing it to a causal seasonal emission mechanism, we propose a month encoder that maps each observation month to a learnable temporal context embedding. This embedding provides a coarse statistical cue about recurring month-level conditions associated with the training labels.

Given a month index $m \in \{1, 2, ..., 12\}$, we first normalize it to the range $[0, 1]$ and convert it into a radian angle $\theta = 2\pi \cdot \frac{m - 1}{12}$ to reflect its position in the annual cycle. To model continuous temporal transitions, we adopt a multi-frequency sinusoidal encoding strategy inspired by positional encodings in Transformers \cite{Vaswani_2017}. Specifically, for a set of $n$ exponentially scaled base frequencies $\{f_i\}_{i=1}^n$, defined as follows:
\begin{equation}
f_i = \exp\left(\frac{i - 1}{n - 1} \cdot \gamma\right), \quad i = 1, 2, ..., n
\end{equation}
where $n$ represents the number of base frequencies, and $\gamma$ is a hyperparameter that controls the maximum log-frequency (i.e., $f_n = \exp(\gamma)$).

We compute the sine and cosine of each frequency-scaled angle and concatenate the results to form the base temporal feature vector:
\begin{equation}
\textbf{v}_{\text{base}} = \left[\sin(f_1 \theta), \cos(f_1 \theta), \ldots, \sin(f_n \theta), \cos(f_n \theta)\right]
\end{equation}

This representation provides a multi-frequency encoding of the month index. To project the features into a temporal embedding space, we apply a non-linear transformation followed by normalization:
\begin{equation}
\textbf{e}_{\text{base}} = \text{LayerNorm}(\text{ReLU}(\textbf{W} \cdot \textbf{v}_{\text{base}} + \textbf{b}))
\end{equation}
where $\textbf{W}$ and $\textbf{b}$ are learnable parameters.

While $\textbf{e}_{\text{base}}$ provides a continuous encoding of month position, it may not fully reflect coarser temporal groups that recur within a year. We therefore incorporate a learnable coarse calendar-group embedding. For implementation, the 12 months are partitioned into predefined coarse temporal groups, and each group is represented using a learnable embedding from a trainable embedding table:
\begin{equation}
\textbf{e}_{\text{group}} = \text{Embedding}[g(m)]
\end{equation}
where Embedding is a trainable embedding table that maps each group index to a corresponding $d$-dimensional embedding vector. The index $g(m)$ is determined based on the given month $m$. In our experiments, Beijing uses four calendar groups and Singapore uses two rainy/dry groups, which serve as coarse temporal partitions rather than causal emission categories.

To achieve a smooth blend between fine-grained and coarse-grained temporal features, we define a cosine-based transition weight:
\begin{equation}
\alpha = 0.5 \cdot |\cos(T\theta - \phi)|
\end{equation}
where $T$ controls the periodicity, $\phi$ denotes the phase shift, and $|\cdot|$ represents the absolute value. By adjusting $T$ and $\phi$, we can flexibly align the transition curve with city-specific temporal groupings, allowing the model to encode month-level context in a continuous manner.

The final month embedding is then computed by weighted interpolation:
\begin{equation}
\textbf{e}_{\text{month}} = (1 - \alpha) \cdot \textbf{e}_{\text{base}} + \alpha \cdot \textbf{e}_{\text{group}}
\end{equation}

This design allows the month encoder to represent both continuous month position and coarser calendar-group context in a unified embedding.

\subsection{Dual-Branch Contrastive Pretraining}
To align modality-specific representations from satellite imagery, street-view textual descriptions, and temporal information, CarbonCLIP adopts two separate contrastive learning strategies for spatial-semantic and temporal alignment. This decoupled design avoids manual tuning of loss weights and reduces conflicts between modality-specific objectives, enabling the model to learn complementary spatial-temporal representations more effectively.

\subsubsection{Image-Text Contrastive Alignment}

To enforce semantic consistency between satellite imagery and their corresponding street-view textual descriptions, we adopt a symmetric contrastive learning objective inspired by contrastive language-image pretraining (CLIP) \cite{Radford_2021}. Specifically, we utilize a visual encoder (e.g., ViT \cite{Dosovitskiy_2020}) followed by a trainable projection layer to extract satellite image spatial embeddings. Meanwhile, descriptive captions are generated from street-view images using LMMs. These captions are further transformed into feature embeddings through a separate trainable projection layer, enabling both modalities to be mapped into a shared embedding space suitable for contrastive learning.

For each image-text pair $(\mathbf{I}_i^{s}, \mathbf{T}_i)$ in a mini-batch of size $N$, where $\mathbf{I}_i^{s}$ denotes the spatial feature embedding of the $i$-th satellite image and $\mathbf{T}_i$ denotes its corresponding tile-level street-view semantic embedding, we contrast the aligned pair against all mismatched pairs within the batch. This encourages the model to maximize the mutual information between positive image-text pairs while minimizing it for negative pairs.

We define a global image-text contrastive loss $\mathcal{L}_{IT}$ as the average of the two symmetric objectives:
\begin{equation}
\label{eq:it_loss}
\begin{aligned}
\mathcal{L}_{IT} = -\frac{1}{2N}  \Bigg[ &\sum_{i}^{N}
\log \frac{\exp\left( S(\mathbf{I}_i^{s}, \mathbf{T}_i) / \tau \right)}{
\sum_{j}^{N} \exp\left( S(\mathbf{I}_i^{s}, \mathbf{T}_j) / \tau \right)} \\
+ &\sum_{i}^{N}
\log \frac{\exp\left( S(\mathbf{T}_i, \mathbf{I}_i^{s}) / \tau \right)}{
\sum_{j}^{N} \exp\left( S(\mathbf{T}_i, \mathbf{I}_j^{s}) / \tau \right)} \Bigg]
\end{aligned}
\end{equation}
where $S(\cdot, \cdot)$ denotes the cosine similarity between normalized embeddings, and $\tau$ is a temperature parameter that scales the logits to control the sharpness of the probability distribution. The loss encourages each satellite image to be most similar to its corresponding street-view description and vice versa, forming a bidirectional alignment in the shared embedding space.

This symmetric contrastive formulation improves multimodal fusion by leveraging both image-to-text and text-to-image associations, which is critical for learning robust joint representations in carbon emission prediction tasks.

\subsubsection{Temporal Alignment with Month Embeddings}

To incorporate temporal awareness into the learned representations, we develop a symmetric contrastive alignment strategy between satellite image temporal features and their corresponding month embeddings. This alignment helps CarbonCLIP learn temporally smooth and cyclically aware representations that are beneficial for tasks with recurrent month-level variation.

Given a mini-batch of size $N$, let $\mathbf{I}_i^{t}$ denote the temporal projection of the $i$-th satellite image feature, and $\mathbf{M}_i$ denote the corresponding embedding from the month encoder based on its month index $m_i \in \{1, 2, \dots, 12\}$.

To softly reflect the cyclic temporal structure of months, we define the circular month distance of month $m_i$ and $m_j$ as $\Delta_{ij} = \min(|m_i - m_j|,\ 12 - |m_i - m_j|)$ and convert it into a soft alignment weight using a Gaussian kernel:
\begin{equation}
w_{ij} = \exp\left( - \frac{ \Delta_{ij}^2 }{ 2\sigma^2 } \right)
\end{equation}
where $\sigma$ is a hyperparameter to determine the width of the similarity window. By adjusting $\sigma$, the framework can be adapted to different month-level temporal similarity patterns. The final temporal alignment loss is the average of two symmetric soft contrastive objectives:
\begin{equation}
\label{eq:tm_loss}
\begin{aligned}
\mathcal{L}_{IM} = -\frac{1}{2N} \sum_{i}^{N} &\Bigg[
\sum_{j}^{N} w_{ij} \cdot \log  \frac{\exp\left( S(\mathbf{I}_i^{t}, \mathbf{M}_j) / \tau \right)}{
    \sum_{k}^{N} \exp\left( S(\mathbf{I}_i^{t}, \mathbf{M}_k) / \tau \right)} \\
+ &
\sum_{j}^{N} w_{ij} \cdot \log \frac{\exp\left( S(\mathbf{M}_i, \mathbf{I}_j^{t}) / \tau \right)}{
    \sum_{k}^{N} \exp\left( S(\mathbf{M}_i, \mathbf{I}_k^{t}) / \tau \right)} \Bigg]
\end{aligned}
\end{equation}
where $S(\cdot, \cdot)$ denotes the cosine similarity between normalized embeddings, and $\tau$ is a temperature parameter that scales the logits to control the sharpness of the probability distribution.

This symmetric contrastive formulation encourages temporally similar satellite images to lie closer to their corresponding month embeddings in the shared space, while cyclic month distance ensures smooth alignment across year-end boundaries (e.g., December and January). By softly weighting the alignment based on temporal proximity, the model learns temporally aware representations that are useful for monthly environmental prediction tasks.

\subsection{Satellite-Only Carbon Emission Predictor}

To evaluate the effectiveness of the learned satellite image representations for real-world downstream tasks, we design a carbon emission prediction module that estimates monthly carbon emissions at the region level. In this phase, we freeze the vision encoder and feature projection layers pretrained on image-text and temporal alignment tasks, and train a lightweight MLP predictor with satellite image and carbon emission data pairs.

For each satellite image tile, we extract two types of features using the pretrained model: a spatial embedding that captures spatial semantics and urban morphology, and a temporal embedding that encodes month-level temporal context. These features are generated by feeding satellite images into a frozen vision encoder followed by two separate projection layers, a spatial projection and a temporal projection. The outputs are normalized and concatenated to form the final feature vector:
\begin{equation}
\mathbf{z}_i = \left[ \text{Norm}\left(\mathbf{I}^{s}_i\right) \,\|\, \text{Norm}\left(\mathbf{I}^{t}_i\right) \right]
\end{equation}
where $\mathbf{I}^{s}_i$ and $\mathbf{I}^{t}_i$ denote the spatial and temporal feature embeddings of the $i$-th image, and $\|$ indicates feature concatenation. Then, we use a simple MLP as the carbon emission predictor, which maps the concatenated feature vector $\mathbf{z}_i$ to a scalar emission value $\hat{y}_i$:
\begin{equation}
\hat{y}_i = f_{\theta}(\mathbf{z}_i)
\end{equation}
where $f_{\theta}(\cdot)$ denotes the learnable MLP with ReLU activations. This design enables fast convergence and avoids overfitting, as the upstream encoders are already well-trained.

The predictor is trained to minimize the mean squared error (MSE) loss between the predicted and ground-truth emissions:
\begin{equation}
\mathcal{L}_{\text{reg}} = \frac{1}{NM} \sum_{i=1}^{N} \sum_{m=1}^{M} (y_{i,m} - \hat{y}_{i,m})^2
\end{equation}
where $y_{i,m}$ denotes the ground-truth carbon emission for the corresponding region $i$ and month $m$. During training, each emission label is divided by a fixed constant of 1000 for numerical stability.

This predictor serves as a carbon emission estimation validation step for our multimodal pretraining framework, demonstrating that the learned image features encode task-relevant spatial-temporal knowledge for urban-scale carbon emission estimation.

\section{Performance Evaluation}
\subsection{Experimental Setup}
\subsubsection{Dataset}
\begin{table}[t]
    \centering
    \caption{Dataset details.}
    \setlength{\tabcolsep}{3pt}
    \begin{tabular}{
        >{\raggedright\arraybackslash}m{2.4cm} 
        >{\raggedright\arraybackslash}m{2.6cm}
        >{\centering\arraybackslash}m{2.35cm}
    }
    \toprule
    \textbf{Data} & \textbf{Source} & \makecell{\textbf{Resolution /}\\\textbf{Spatial granularity}}\\
    \midrule
    Satellite Imagery & Planet \cite{planet_api} & 3~m\\
    Street-View Imagery & Singapore \cite{googlemaps} / Beijing \cite{baidumap} & 200~m\\
    Carbon Emissions & ODIAC \cite{Oda_2018, odiac2023} & 1~km\\
    \bottomrule
    \end{tabular}
    
    \label{tab:dataset}
\end{table}

The datasets used in this study comprise satellite imagery, street-view imagery, and carbon emission data for two representative Asian cities with different climatic and temporal regimes: Beijing and Singapore (see Table~\ref{tab:dataset} for detailed information).

We utilize high-resolution satellite imagery provided by Planet \cite{planet_api} for constructing monthly satellite observations in both Beijing and Singapore, which provides high-resolution imagery at a resolution of 3 meters per pixel. Both the satellite imagery and carbon emission data used in this study correspond to the year 2022. To ensure temporal consistency and minimize the influence of atmospheric conditions, we manually select one representative image for each month by using online filtering and preview tools in the Planet platform. Specifically, for each city and month, we prioritize images with less than ten percent cloud coverage, verified via both automated cloud masks and manual inspection of RGB composites. This selection strategy balances image availability and quality, allowing us to observe monthly surface changes with minimal cloud artifacts. For Singapore, no suitable September image with sufficiently low cloud coverage could be identified because of frequent tropical cloud cover; therefore, September was excluded from the Singapore dataset to maintain data quality. To align with the spatial resolution of the emission dataset and facilitate regional analysis, we aggregate and clip the Planet imagery into non-overlapping 1~km\textsuperscript{2} spatial tiles. After monthly quality filtering, the remaining satellite image-label pairs form the candidate pool before applying street-view coverage constraints.

Street-view imagery was systematically collected to provide rich ground-level semantic context for multimodal urban representation learning. Due to regional accessibility constraints, imagery was obtained from two platforms: Google Maps Street View for Singapore~\cite{googlemaps} and Baidu Maps Street View for Beijing~\cite{baidumap}. To ensure comprehensive, spatially uniform, and semantically diverse coverage, while avoiding biases introduced by population density or road hierarchy, we adopted a fixed-interval sampling strategy based on road network geometry. Specifically, we extracted complete drivable road networks for both cities from OpenStreetMap (OSM). Along each road segment, virtual sampling points were placed at regular 200-meter intervals. This step size was empirically chosen to balance wide geographic coverage with redundancy control, particularly in areas with dense road meshes. The strategy ensures equitable sampling across diverse urban forms, including central business districts, residential neighborhoods, and industrial zones. At each sampling point, the nearest available panoramic street-view image was queried using the corresponding API. In total, this process yielded 45,408 street-view images in Beijing and 20,415 images in Singapore. To fuse street-level semantics with satellite imagery, each street-view image was mapped to the 1~km\textsuperscript{2} spatial tile containing its GPS coordinates. After this spatial matching step, 45,398 panoramas in Beijing and 15,976 panoramas in Singapore were retained for tile-level semantic representation. Because street-view coverage is spatially static in this study, the spatial tile counts are computed once per city rather than expanded into month-level satellite observations. The resulting cross-modal linkage enables robust semantic enrichment and supervision during multimodal pretraining, with street-view imagery subsequently used as input to Qwen2.5-VL~\cite{Bai_2025} for dense text generation.

\begin{table}[H]
    \centering
    \caption{Street-view coverage over 1~km\textsuperscript{2} spatial tiles counted once per city.}
    \label{tab:streetview_tile_statistics_targeted}
    \setlength{\tabcolsep}{5pt}
    \begin{tabular}{l r r}
        \toprule
        \textbf{Statistic} & \textbf{Beijing} & \textbf{Singapore} \\
        \midrule
        1~km\textsuperscript{2} spatial tiles & 1,021 & 512 \\
        Matched panoramas & 45,398 & 15,854 \\
        Panoramas/tile (mean) & 44.46 & 30.96 \\
        Panoramas/tile (median) & 42 & 29.0 \\
        \bottomrule
    \end{tabular}
\end{table}

The statistics in Table~\ref{tab:streetview_tile_statistics_targeted} show that most tile-level semantic anchors are supported by multiple panoramas, which motivates constructing tile-level semantic anchors from the matched panoramas rather than treating any single panorama as the complete semantic label of a 1~km\textsuperscript{2} area.

We use monthly gridded carbon dioxide emission estimates from the Open-source Data Inventory for Anthropogenic CO$_2$ (ODIAC) dataset~\cite{Oda_2018, odiac2023}. ODIAC provides globally consistent fossil-fuel carbon dioxide emissions at a high spatial resolution of 1~km\textsuperscript{2} and a monthly temporal frequency. Owing to its fine granularity and observation-constrained construction, ODIAC has been widely adopted for urban-scale carbon monitoring, emission estimation, and policy evaluation~\cite{Yan_2024, Tian_2024, Mun_2025, Hao_2025}. To construct image-label pairs, each study region was partitioned into non-overlapping 1~km\textsuperscript{2} tiles consistent with the spatial resolution of both the satellite imagery and the ODIAC grids. For every tile and each month, a corresponding satellite image was independently extracted and paired with the exact ODIAC carbon emission value for the same spatial tile and temporal interval, forming a one-to-one image-label correspondence. To ensure strict temporal consistency and high-quality supervision, we adopted a conservative quality-control strategy. If a satellite image corresponding to a given tile-month pair was missing or deemed invalid due to cloud contamination, sensor noise, or other acquisition issues, the associated carbon emission label for that month was discarded. This filtering step inevitably leads to variability in the number of valid samples across months and locations.

\subsubsection{Model Settings}
We detail the key model components and configurations used, including the street-view description generator and the multimodal encoders.
\begin{itemize}
    \item \textbf{Street-View Description Generator}. 
    To balance semantic fidelity and computational efficiency, we adopt the Qwen2.5-VL-7B model to generate fine-grained street-view descriptions. Table~\ref{tab:qualitative_comparison} provides a qualitative comparison of descriptions generated by Qwen2.5-VL models with different parameter sizes. As a small-scale generator-selection check, we manually score twenty held-out panoramic street-view images, ten from each city, under the same prompt. Each generated description is rated on a 1--10 scale according to three criteria: factual grounding, which measures whether visual claims are supported by the image; carbon relevance, which measures whether the description focuses on emission-related urban factors such as road infrastructure, land use, building density, traffic facilities, greenery, and industrial or commercial activity; and hallucination-free quality, which measures whether unsupported visual claims are avoided. As summarized in Table~\ref{tab:text_quality_targeted}, Qwen2.5-VL-7B improves over the 3B model in factual grounding and hallucination control, while remaining close to the 72B model across all three criteria. We therefore choose the 7B model because it provides a favorable tradeoff between semantic quality and computational cost for large-scale street-view description generation.
    
    \item \textbf{Image and Text Encoders}. 
    For contrastive pretraining, we employ a ViT-B/32 architecture as the satellite image encoder and a Transformer-based architecture as the text encoder. Both encoders are initialized with pretrained weights from the official \texttt{openai/clip-vit-base-patch32} model~\cite{Radford_2021}.
\end{itemize}

\begin{table}[h]
    \caption{Comparison of generated descriptions from Qwen2.5-VL models with different parameter sizes (3B, 7B, and 72B) given the same panoramic street-view image and prompt. We highlight accurate details, factual errors, and redundant information to illustrate the impact of model scaling on semantic quality.}
    \label{tab:qualitative_comparison}
    \centering
    \scalebox{1}{
    \begin{tabular}{c|m{14.7cm}}
        \toprule
        \multicolumn{2}{c}{\includegraphics[width=0.5\textwidth]{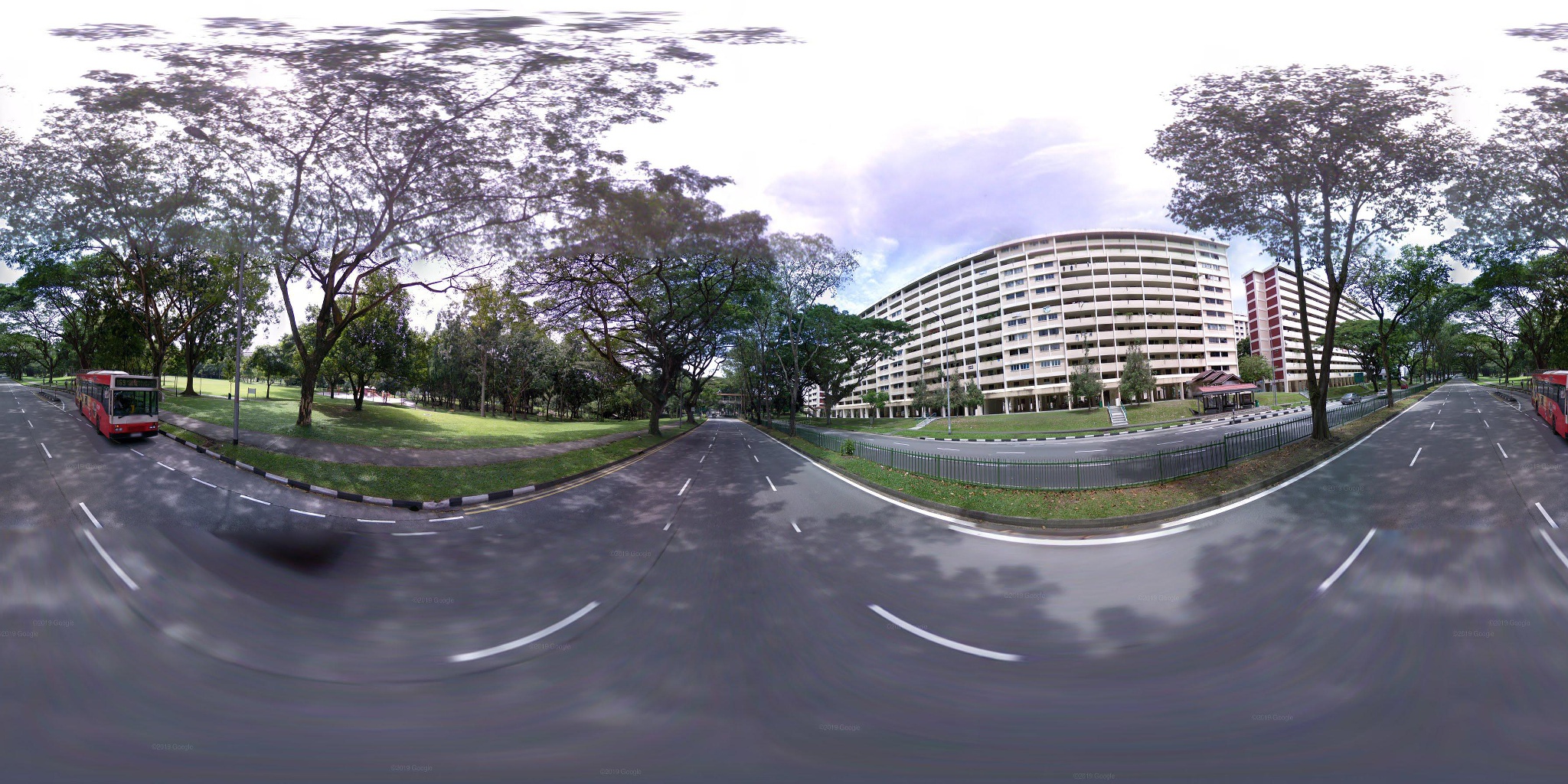}} \\  \midrule
        \textbf{Model Size} & \textbf{Generated Description} \\ \midrule
        3B & The panoramic street-view image of Singapore displays an urban scene featuring \textcolor{green}{a large apartment building with multiple balconies} arranged in a curved profile, \textcolor{green}{bordered by trees with dense foliage} on the left side and \textcolor{red}{a solid fence on the right}. This building suggests a residential area. The \textcolor{green}{two-way roads} curving around the buildings are \textcolor{green}{lined with green lawns and spaces} where sidewalk pedestrian traffic is expected. \textcolor{red}{Administrative structures are likely within or adjacent to this building complex.} There is also \textcolor{green}{some public transport} visible as red buses parked in designated lanes. The presence of fenceline and open space indicates provisions for both residential and commuter-friendly infrastructure. \\ \midrule
        7B & The panoramic image showcases an urban environment with \textcolor{green}{a large multi-story residential building} prominently situated in the center-right portion of the frame. The building exhibits a uniform architectural style with \textcolor{green}{multiple balconies and windows}, suggesting a planned residential area. Surrounding the building are \textcolor{green}{lush green trees} lining both sides of a wide road, indicating a well-maintained urban landscape. \textcolor{green}{The road appears to be divided into lanes}, with a median strip separating the traffic flow. On the left side of the image, there is a red bus partially visible, hinting at \textcolor{green}{public transportation services available} in this area. The overall scene reflects a clean and organized urban setting, likely part of a planned community or estate in Singapore. \\ \midrule
        72B & The panoramic street-view image at \textcolor{blue}{coordinates (103.833802, 1.367105)} showcases a well-organized urban environment in Singapore characterized by \textcolor{green}{a wide, multi-lane road} flanked by \textcolor{green}{lush greenery and mature trees} providing ample shade. On either side of the road, there are sidewalks bordered by \textcolor{blue}{black and white striped curbs}, with a bus stop visible on the right side indicating \textcolor{green}{public transportation accessibility}. The presence of \textcolor{green}{a large, curved residential building} with multiple stories suggests a high-density living area designed for efficiency and space optimization. Adjacent to the building, there is an open grassy area with scattered trees, possibly \textcolor{green}{serving as a communal park or recreational space}. The overall layout reflects a balance between urban infrastructure and natural elements, \textcolor{blue}{emphasizing pedestrian-friendly pathways and green spaces integrated within the built environment}. The image also captures another bus on the left side of the road, reinforcing the area's connectivity and reliance on public transit. The architectural style of the buildings and the structured urban planning suggest a modern and planned cityscape. \\ \bottomrule
\end{tabular}}
\end{table}

\begin{table}[!t]
    \centering
    \caption{Small-scale manual quality assessment of generated street-view descriptions of 20 samples (10 from each city). Scores range from 1 to 10, and higher scores are better for all metrics.}
    \label{tab:text_quality_targeted}
    \setlength{\tabcolsep}{3pt}
    \begin{tabular}{l c c c}
        \toprule
        \textbf{Model} & \makecell{\textbf{Factual}\\\textbf{Grounding}} & \makecell{\textbf{Carbon}\\\textbf{Relevance}} & \makecell{\textbf{Hallucination-}\\\textbf{free}} \\
        \midrule
        Qwen2.5-VL-3B & 7.1 & 7.1 & 7.0 \\
        Qwen2.5-VL-7B & 8.4 & 7.8 & 7.9 \\
        Qwen2.5-VL-72B & 8.5 & 8.2 & 8.0 \\
        \bottomrule
    \end{tabular}
\end{table}

\subsubsection{Baselines}
We compare the performance of CarbonCLIP with the following baselines in carbon emission prediction:
\begin{itemize}
    \item \textbf{ResNet} \cite{He_2016}. ResNet is a classic and widely used convolutional neural network architecture composed of residual blocks. We adopt a ResNet-18 model pretrained on ImageNet and fine-tune it on the carbon emission prediction task using satellite imagery as input. This serves as a strong convolutional baseline to evaluate the effectiveness of spatial feature extraction.
    \item \textbf{Vision Transformer (ViT)} \cite{Dosovitskiy_2020}. ViT is a transformer-based architecture that divides an image into patches and applies self-attention mechanisms to model global dependencies across them, offering a powerful alternative to convolutional networks for visual tasks. In our baseline, we use ViT to extract spatial representations from satellite images and train a lightweight regression head to predict carbon emissions. This allows us to evaluate the effectiveness of transformer-based spatial encoding under a purely image-based setting.
    \item \textbf{UrbanCLIP} \cite{Yan_2024}. UrbanCLIP is a multimodal urban representation model that generates textual descriptions from satellite images and enhances visual features through CLIP-style contrastive learning. In our adaptation, we follow its unimodal inference setup by using only satellite images to extract pretrained embeddings, which are then fed into a regression head for carbon emission prediction.
\end{itemize}

All methods are evaluated under a controlled protocol using the same train, validation, and test partitions, ODIAC emission labels, fixed label-scaling procedure, and evaluation metrics. After each representation is obtained, the downstream carbon-emission regressor is trained using the same downstream training and evaluation procedure. This design keeps the comparison focused on the learned representation rather than differences in data preparation or evaluation settings. All downstream regressors use the same three-layer regression head, with 512 neurons in the first hidden layer, 256 neurons in the second hidden layer, and a final scalar output layer. We do not set a fixed number of training epochs; instead, each model is trained with early stopping and terminated when the validation performance does not improve for the most recent 10 validation steps. Therefore, the exact number of epochs can vary across methods and runs. For UrbanCLIP, we follow its satellite-only inference setting by extracting pretrained satellite embeddings and feeding them into the same downstream regressor. No baseline receives street-view imagery, street-view text, or month embeddings during inference. Table~\ref{tab:fair_comparison_targeted} summarizes the encoder, auxiliary supervision used during representation learning, and test-time input.

\begin{table}[!t]
    \centering
    \caption{Controlled evaluation settings for baselines and CarbonCLIP. All methods use the same split, emission labels, fixed label scaling, and metrics.}
    \label{tab:fair_comparison_targeted}
    \setlength{\tabcolsep}{2pt}
    \renewcommand{\arraystretch}{1.12}
    \begin{tabular}{
        >{\raggedright\arraybackslash}p{1.55cm}
        >{\raggedright\arraybackslash}p{2.05cm}
        >{\raggedright\arraybackslash}p{1.95cm}
        >{\raggedright\arraybackslash}p{2.10cm}
    }
        \toprule
        \textbf{Method} & \textbf{Backbone} & \textbf{Aux. Supervision} & \textbf{Test-time Input} \\
        \midrule
        ResNet & ResNet-18 & None & Satellite tile \\
        ViT & ViT-B/32 & None & Satellite tile \\
        UrbanCLIP & CoCa-ViT-L/14 & Satellite-derived text & Satellite embedding \\
        CarbonCLIP & ViT-B/32 & Street-view text and month context & Satellite embedding \\
        \bottomrule
    \end{tabular}
\end{table}

\begin{table}[!htbp]
    \caption{Performance comparison of different methods on Beijing and Singapore across seasons in 2022. The best results are in \textbf{bold} and the second-best results are \underline{underlined}. All results are averaged over three runs, with CarbonCLIP exhibiting consistently strong performance across both cities.}
    \setlength{\tabcolsep}{10pt}
    \centering
    \resizebox{\textwidth}{!}{
    \begin{tabular}{c|c|ccccc|ccc}
    \toprule
    \multirow{2}{*}{Method} & \multirow{2}{*}{Metric} & \multicolumn{5}{c|}{Beijing} & \multicolumn{3}{c}{Singapore} \\
    \cmidrule(lr){3-7}\cmidrule(lr){8-10}
     &  & Spring & Summer & Autumn & Winter & Year & Rainy & Dry & Year \\
    \midrule

    \multirow{3}{*}{ResNet~\cite{He_2016}} 
    & $\text{R}^2$$\uparrow$ & 0.451 & 0.356 & 0.168 & 0.474 & 0.386 & \underline{0.575} & 0.552 & 0.569 \\
    & RMSE$\downarrow$ & 0.843 & 1.164 & 1.126 & 1.056 & 1.053 & \underline{0.555} & 0.650 & 0.599 \\
    & MAE$\downarrow$ & 0.657 & 0.879 & 0.940 & 0.827 & 0.824 & \underline{0.431} & 0.503 & \underline{0.463} \\
    \midrule

    \multirow{3}{*}{ViT~\cite{Dosovitskiy_2020}} 
    & $\text{R}^2$$\uparrow$ & 0.463 & \underline{0.588} & 0.488 & 0.604 & 0.559 & 0.517 & 0.558 & 0.544 \\
    & RMSE$\downarrow$ & 0.834 & \underline{0.931} & 0.883 & 0.917 & 0.892 & 0.591 & 0.645 & 0.616 \\
    & MAE$\downarrow$ & 0.601 & \underline{0.660} & 0.661 & 0.671 & 0.648 & 0.469 & 0.497 & 0.482 \\
    \midrule

    \multirow{3}{*}{UrbanCLIP~\cite{Yan_2024}} 
    & $\text{R}^2$$\uparrow$ & \underline{0.628} & 0.582 & \underline{0.703} & \underline{0.640} & \underline{0.642} & 0.537 & \underline{0.602} & \underline{0.576} \\
    & RMSE$\downarrow$ & \underline{0.694} & 0.938 & \underline{0.672} & \underline{0.873} & \underline{0.803} & 0.579 & \underline{0.613} & \underline{0.594} \\
    & MAE$\downarrow$ & \underline{0.540} & 0.674 & \underline{0.510} & \underline{0.617} & \underline{0.586} & 0.460 & \underline{0.469} & 0.464 \\
    \midrule

    \multirow{3}{*}{CarbonCLIP (Proposed)} 
    & $\text{R}^2$$\uparrow$ & \textbf{0.716} & \textbf{0.696} & \textbf{0.737} & \textbf{0.737} & \textbf{0.728} & \textbf{0.687} & \textbf{0.712} & \textbf{0.704} \\
    & RMSE$\downarrow$ & \textbf{0.606} & \textbf{0.800} & \textbf{0.632} & \textbf{0.747} & \textbf{0.701} & \textbf{0.476} & \textbf{0.521} & \textbf{0.496} \\
    & MAE$\downarrow$ & \textbf{0.449} & \textbf{0.580} & \textbf{0.478} & \textbf{0.550} & \textbf{0.514} & \textbf{0.349} & \textbf{0.379} & \textbf{0.362} \\

    \bottomrule
    \end{tabular}
    }
    
    \label{tab:results}
\end{table}

\subsubsection{Evaluation Metrics} To evaluate prediction performance, we adopt three widely used metrics: the coefficient of determination (R\textsuperscript{2}), root mean squared error (RMSE), and mean absolute error (MAE). A higher R\textsuperscript{2} and lower RMSE and MAE indicate better predictive accuracy.

\begin{figure}[!htbp]
    \centering
    \includegraphics[width=\columnwidth]{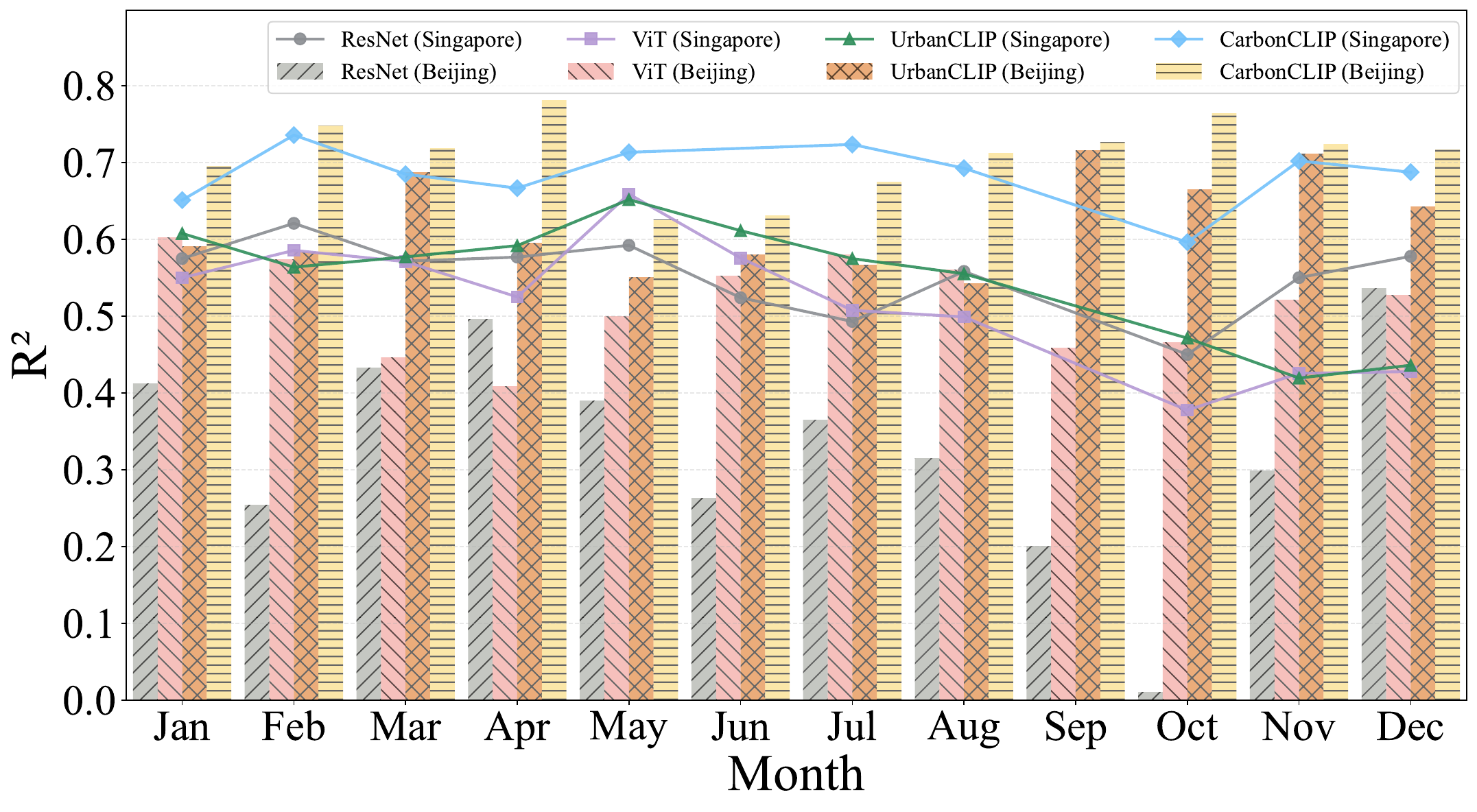}
    \caption{Monthly carbon emission R\textsuperscript{2} comparison of different methods on Beijing and Singapore.}
    \label{fig:cities_r2_comparison}
\end{figure}

\subsection{Performance Evaluation}

To evaluate CarbonCLIP, we compare it with representative baselines on the collected datasets. Table~\ref{tab:results} and Fig.~\ref{fig:cities_r2_comparison} present the overall results, from which we obtain the following three main findings:\\
\textbf{1) CarbonCLIP achieves the best overall performance and temporal consistency across both cities.}
As reported in Table~\ref{tab:results}, CarbonCLIP consistently outperforms all baseline methods in terms of R\textsuperscript{2}, RMSE, and MAE for both Beijing and Singapore. In Beijing, CarbonCLIP attains an annual R\textsuperscript{2} of 0.728, substantially exceeding UrbanCLIP (0.642), ViT (0.559), and ResNet (0.386). Likewise, in Singapore, CarbonCLIP achieves an R\textsuperscript{2} of 0.704, outperforming UrbanCLIP (0.576), ViT (0.544), and ResNet (0.569). The baseline models exhibit relatively similar performance in Singapore, suggesting that existing approaches may share limitations in feature extraction. These results indicate that CarbonCLIP provides a stronger satellite-based representation for monthly urban carbon emission prediction.\\
\textbf{2) CarbonCLIP consistently achieves top performance throughout the year, showing stable monthly prediction.}
As shown in Fig.~\ref{fig:cities_r2_comparison}, CarbonCLIP maintains stronger monthly performance than the baselines across Beijing and the available Singapore observations. ResNet and ViT suffer from noticeable degradation during several monthly periods, while UrbanCLIP is more stable but still fluctuates. In contrast, CarbonCLIP achieves consistently high R\textsuperscript{2} values with lower variation, reflecting more stable monthly prediction under the satellite-only inference setting.\\
\textbf{3) These results highlight the benefit of fusing spatial and temporal priors through contrastive pretraining.}
The advantage of CarbonCLIP stems from its ability to integrate spatial semantics and month-level temporal context by aligning satellite features with street-view-generated text and temporal embeddings during pretraining. Therefore, without explicitly using street-view imagery at inference, CarbonCLIP learns satellite representations that are semantically richer than traditional CNN or transformer backbones.

\subsection{Ablation Studies}

\begin{figure}[!htbp]
    \centering
    \includegraphics[width=\columnwidth]{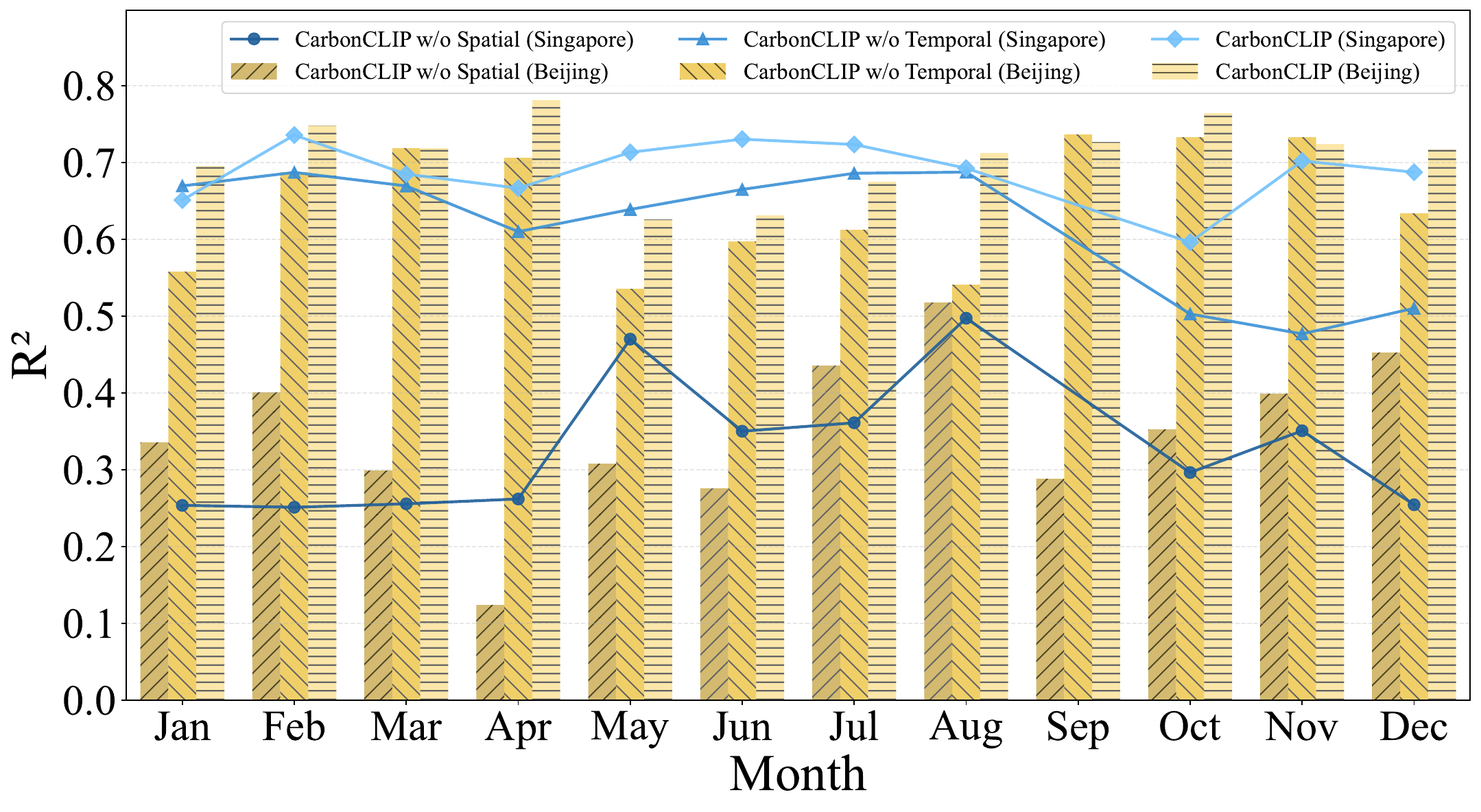}
    \caption{Ablation studies on Beijing and Singapore.}
    \label{fig:cities_ablation_comparison}
\end{figure}

As shown in Fig.~\ref{fig:cities_ablation_comparison}, we first assess the contribution of two key components in our CarbonCLIP framework: spatial embedding and temporal embedding. We compare CarbonCLIP with its two ablated variants, one without the temporal branch and one without the spatial branch, by examining the monthly R\textsuperscript{2} performance across the year. We further include a targeted ablation on street-view supervision to test whether direct street-view image alignment is sufficient or whether LMM-generated semantic abstraction provides stronger supervision.

\subsubsection{Effect of Spatial Embedding}
Removing the spatial branch leads to significant performance degradation across all months. This suggests that detailed spatial semantics such as land-use types, building distribution, and infrastructure density, captured by the spatial projection layer, are essential for accurately estimating urban carbon emissions. Without the spatial branch, CarbonCLIP lacks the ability to ground urban carbon emissions in the broader urban context.
\subsubsection{Effect of Temporal Embedding}
Excluding the temporal branch also reduces performance, though less severely than removing the spatial branch. The gap is especially noticeable in months where emission labels are associated with recurrent energy-use and activity patterns. The temporal projection allows CarbonCLIP to align image features with explicitly trained month embeddings, helping encode month-level contextual correlations that static spatial features alone cannot represent.

\subsubsection{Street-View Image Alignment Ablation}
\label{sec:text_visual_ablation_targeted}
We further conduct a targeted downstream ablation to compare direct street-view image alignment with the proposed CarbonCLIP design. In the street-view image-alignment variant, we replace CarbonCLIP's spatial semantic alignment target with raw street-view visual embeddings. Specifically, each matched street-view panorama is encoded by the CLIP image encoder, and all panorama embeddings within the same 1~km\textsuperscript{2} spatial tile are aggregated by mean pooling:

\begin{equation}
    \mathbf{V}_i = \frac{1}{K_i} \sum_{k=1}^{K_i} \mathbf{v}_{i,k},
    \label{eq:targeted_visual_aggregation}
\end{equation}

where $\mathbf{v}_{i,k}$ denotes the CLIP image embedding of the $k$-th street-view panorama in tile $i$. The pooled street-view image embedding is then used as the spatial alignment target by substituting it for the street-view text embedding in Eq.~(\ref{eq:it_loss}). The satellite encoder, temporal branch, training split, and downstream regression head are kept consistent with CarbonCLIP, so the comparison isolates whether direct street-view image alignment is sufficient or whether LMM-generated semantic abstraction provides stronger supervision.

\begin{table}[!t]
    \centering
    \caption{Targeted ablation on street-view supervision. ``SVI Alignment'' denotes direct street-view image alignment.}
    \label{tab:text_visual_ablation_targeted}
    \setlength{\tabcolsep}{4pt}
    \renewcommand{\arraystretch}{1.08}
    \begin{tabular}{l l c c c}
        \toprule
        \textbf{Method} & \textbf{City} & $\textbf{R}^2$$\uparrow$ & \textbf{RMSE}$\downarrow$ & \textbf{MAE}$\downarrow$ \\
        \midrule
        SVI Alignment & Beijing & 0.5396 & 0.9112 & 0.6917 \\
        SVI Alignment & Singapore & 0.5892 & 0.5843 & 0.4558 \\
        \midrule
        CarbonCLIP & Beijing & \textbf{0.728} & \textbf{0.701} & \textbf{0.514} \\
        CarbonCLIP & Singapore & \textbf{0.704} & \textbf{0.496} & \textbf{0.362} \\
        \bottomrule
    \end{tabular}
\end{table}
As shown in Table~\ref{tab:text_visual_ablation_targeted}, CarbonCLIP outperforms the direct street-view image-alignment variant in both cities. One possible reason is that directly using street-view images as the alignment target can be more sensitive to inconsistent panorama-level cues within the same tile, such as different viewing directions, road segments, land-use frontages, or local occlusions. In comparison, generated text may provide a more abstract semantic target by summarizing recurring urban functions from individual views and reducing part of the view-specific noise before satellite alignment.

\section{Discussion}

The experimental results indicate that CarbonCLIP improves monthly carbon emission prediction in both study cities, but the added semantic and temporal signals should be interpreted within the constraints of the current data and task setting. This section discusses three issues: how street-view semantics are distilled into a satellite-only prediction setting, what the learned representation reveals and where it remains limited, and how temporal modeling improves prediction across different climate settings.

\subsection{Semantic Distillation for Satellite-Only Carbon Prediction}

CarbonCLIP differs from general CLIP-style geospatial representation methods mainly in the role assigned to auxiliary modalities. Instead of learning a general location, remote-sensing, or urban representation, CarbonCLIP uses street-view descriptions and month context to shape a satellite representation for monthly carbon emission prediction. The training process is asymmetric: street-view text and month-level context provide supervision during pretraining, while the downstream predictor operates with satellite embeddings. This design keeps inference practical while allowing the satellite encoder to absorb semantic information that is difficult to infer from overhead imagery alone.

The small-scale text-quality assessment and the street-view image-alignment ablation help interpret the source of this gain. Manual scoring in Table~\ref{tab:text_quality_targeted} suggests that, within the sampled cases, the selected Qwen2.5-VL generator produces descriptions with reasonable factual grounding, carbon relevance, and hallucination control. The comparison in Table~\ref{tab:text_visual_ablation_targeted} further shows that replacing text-based supervision with direct street-view image alignment reduces performance in both cities. This result indicates that tile-level street-view heterogeneity is not only a sampling concern but also a practical alignment challenge. One possible explanation is that raw panorama embeddings can be sensitive to view direction, local occlusion, and uneven street-level sampling within the same tile, whereas generated descriptions may provide a more compact semantic target for alignment.

This difference is especially relevant when multiple panoramas are matched to the same satellite tile. Directly pooling street-view image embeddings can retain strong view-specific cues from different road directions, facades, traffic scenes, and local occlusions, so visually heterogeneous panoramas within one 1~km\textsuperscript{2} tile may produce a less coherent alignment target. In contrast, text embeddings derived from generated descriptions can reduce part of this image-level conflict by abstracting repeated street-level observations into functional concepts before tile-level semantic alignment.

\subsection{Learned Representation and Semantic Limits}
\label{sec:interpretability_targeted}

\begin{figure}[!t]
    \centering
    \includegraphics[width=0.95\textwidth]{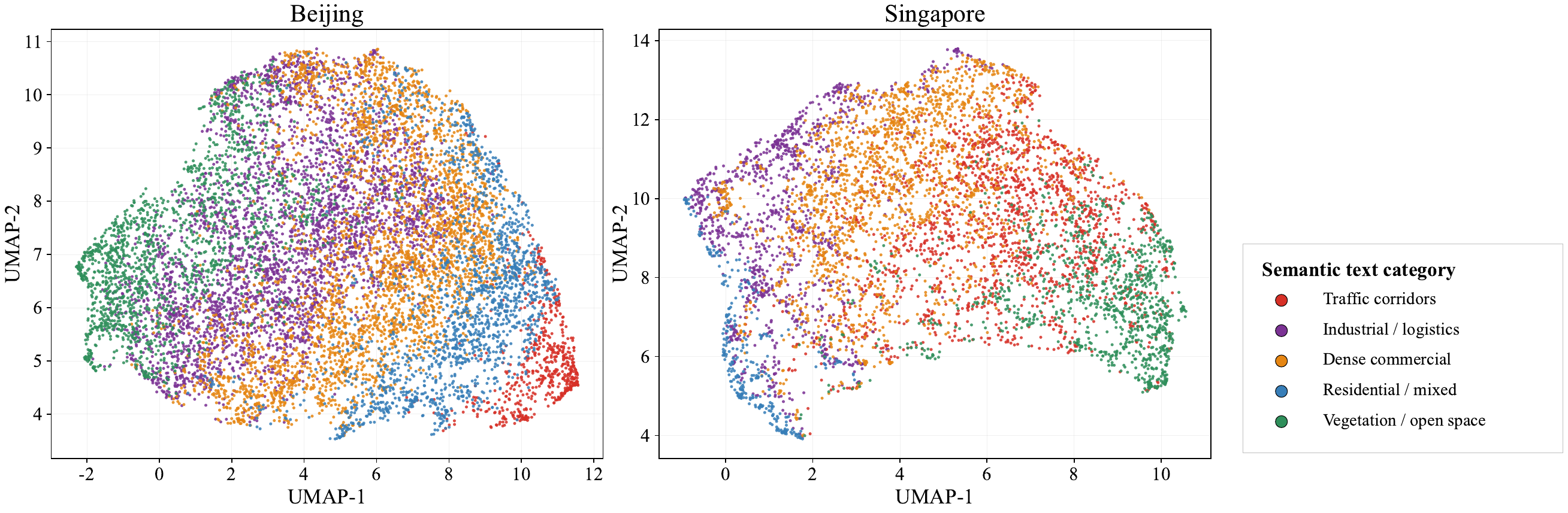}
    \caption{UMAP visualization of CarbonCLIP satellite embeddings. Each point represents a 1~km\textsuperscript{2} spatial tile after averaging valid monthly satellite embeddings. Colors denote the nearest semantic category in the text-embedding space. The Beijing and Singapore panels are generated separately and should be interpreted only for within-city neighborhood structure, not as a cross-city geometric comparison.}
    \label{fig:embedding_visualization_targeted}
\end{figure}

Since street-view text is not used during inference, a central interpretability question is whether multimodal pretraining leaves semantic structure in the satellite embeddings. Fig.~\ref{fig:embedding_visualization_targeted} visualizes the learned satellite embeddings with UMAP and cosine distance. Each point denotes a 1~km\textsuperscript{2} spatial tile, represented by the average satellite embedding across valid monthly observations, and colors indicate the semantic category whose text embedding is closest to the tile-level street-view semantic embedding. Beijing and Singapore are projected with separate UMAP fits so that city-level distributional differences do not dominate either panel. Within each city, the local neighborhoods show that tiles with related urban functions, including traffic corridors, dense commercial areas, residential or mixed areas, and vegetation or open space, tend to occupy coherent regions in the embedding space. This pattern suggests that part of the street-level semantic information introduced during pretraining is reflected in the satellite representation used for inference.

This visualization is qualitative and city-specific, and should not be interpreted as a complete explanation of prediction behavior. The semantic structure can become less distinct when a tile contains mixed land use, when street-view observations are sparse, or when ground-level imagery does not match the satellite observation period. These cases also define the deployment boundary of the current framework. Although inference requires only satellite imagery, pretraining still depends on street-view availability and generated-text quality. Applying CarbonCLIP to cities with limited street-view coverage or substantially different morphology may therefore require additional validation, alternative weak supervision, or stronger self-supervised satellite pretraining.

\subsection{Effectiveness of Temporal Modeling}

Month-level temporal context is important for improving the accuracy and generalizability of monthly urban carbon emission prediction. Our results show that incorporating month embeddings improves performance in both Beijing and Singapore, indicating that the temporal branch provides information that cannot be fully captured by static satellite features. In Beijing, where month-to-month observations are associated with changes in heating and cooling demand, vegetation conditions, and other environmental factors, temporal alignment provides useful contextual variation. In Singapore, although the climate does not follow a typical four-season pattern, rainy and dry periods still introduce temporal differences, and the month embeddings also bring measurable gains. These results suggest that the temporal component is useful across different climate settings.

\begin{center}
\includegraphics[width=\columnwidth]{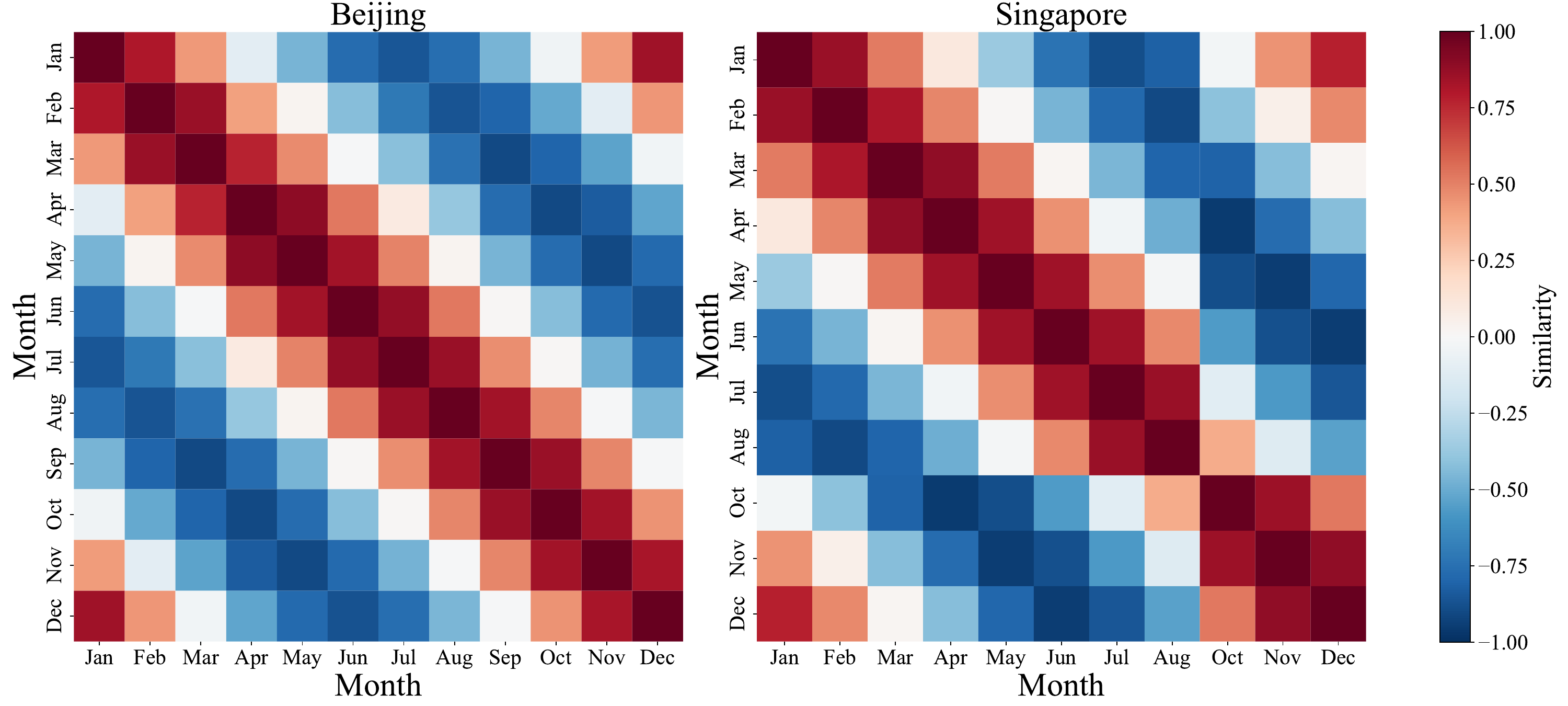}
\captionof{figure}{Similarity of temporal embeddings across months in Beijing (left) and Singapore (right).}
\label{fig:temporal_similarity_comparison}
\end{center}

To further interpret the learned temporal features, Fig.~\ref{fig:temporal_similarity_comparison} visualizes the pairwise similarities among month embeddings for both cities. Both heatmaps exhibit a broadly similar cyclic pattern: adjacent months tend to have higher similarity, whereas months farther apart in the annual cycle tend to be less similar. Within this shared structure, Beijing shows slightly clearer block-like transitions across months, while Singapore presents a more continuous temporal space with several positive off-diagonal month pairs. This comparison suggests that the learned embeddings capture a common cyclic organization while reflecting moderate city-specific temporal variation. The similarity between the two patterns may be related to the limited resolution of 12 discrete month embeddings and the missing September observations in Singapore caused by cloud cover. Finer-grained or continuous temporal representations could further improve the modeling of subtle temporal variations.

\begin{center}
    \begin{minipage}{\columnwidth}
    \centering
    \captionof{table}{Temporal embedding similarity statistics.}
    \label{tab:temporal_embedding_similarity}
    \setlength{\tabcolsep}{5pt}
    \renewcommand{\arraystretch}{1.05}
    \begin{tabular}{lcccc}
        \toprule
        \textbf{City} & \textbf{Adjacent} & \makecell{\textbf{Within}\\\textbf{group}} & \makecell{\textbf{Between}\\\textbf{group}} & \makecell{\textbf{$r$}\\\textbf{(sim., dist.)}} \\
        \midrule
        Singapore & 0.248 & 0.247 & 0.163 & -0.086 \\
        Beijing & 0.490 & 0.605 & -0.065 & -0.658 \\
        \bottomrule
    \end{tabular}
    \end{minipage}
\end{center}

Table~\ref{tab:temporal_embedding_similarity} provides a compact quantitative complement to the visual comparison. Adjacent similarity measures neighboring months under circular month distance, and $r$(sim., dist.) measures the correlation between cosine similarity and circular month distance. Beijing has higher adjacent-month similarity (0.490), higher within-group similarity (0.605), negative between-group similarity (-0.065), and a stronger negative distance correlation ($r=-0.658$), suggesting that its month embeddings form clearer temporal neighborhoods. This pattern is consistent with stronger annual variation in Beijing satellite observations, where vegetation, surface conditions, and energy-use-related context vary more clearly across the year. For Singapore, the tropical rainy/dry regime is expected to vary more gradually in optical satellite observations. In addition, Beijing is represented by four calendar groups, whereas Singapore is represented by two broader rainy/dry groups, so the between-group metric compares different levels of temporal-group granularity. Under this setting, Singapore's lower adjacent-month similarity (0.248), lower within-group similarity (0.247), weaker negative distance correlation ($r=-0.086$), and positive between-group similarity (0.163) are consistent with a more continuous temporal embedding space rather than a sharply separated group structure.

Overall, the temporal analysis supports the design choice of adding a month encoder, but the month index should be understood as a coarse temporal cue rather than a direct explanation of emission drivers. Monthly emissions may also be affected by temperature, heating and cooling demand, traffic activity, electricity use, and other socioeconomic factors that are not separately modeled here. Future work could incorporate consistently available temporal covariates or continuous temporal representations to capture finer-grained variations while preserving the satellite-only inference setting.

\section{Conclusion}
\subsection{Summary}
In this work, we propose CarbonCLIP, a scalable framework designed to overcome the inherent limitations of top-down physical observations in urban carbon emission estimation. By distilling multimodal knowledge into a unified satellite-based representation, our approach enriches static remote sensing imagery with fine-grained human-centric semantics and month-level temporal context. Specifically, we leverage LMMs to automatically generate dense textual descriptions of street-level activities, providing richer semantic information than traditional retrieval-based captions. Simultaneously, we introduce a cyclic month encoding mechanism that allows the model to encode recurrent temporal correlations often missed by vision-only baselines. Crucially, CarbonCLIP adopts an asymmetric learning paradigm: it exploits comprehensive multimodal data during pretraining but requires only satellite imagery for inference. This design supports scalable satellite-only deployment while maintaining the benefits of multimodal awareness. Experiments on Beijing and Singapore validate that CarbonCLIP improves prediction accuracy in both study cities, offering a practical solution for satellite-based urban carbon modeling.

\subsection{Limitations and Future Work}
Although CarbonCLIP achieves strong performance, several limitations remain. First, spatial correlations between neighboring regions are not explicitly modeled. While such context could enhance prediction accuracy, the ODIAC dataset exhibits highly similar spatial patterns across months, limiting the extraction and validation of meaningful inter-regional dependencies. Future work will explore finer-grained or locally validated emission inventories, adaptive spatial units, and uncertainty-aware aggregation to better capture spatial continuity and mixed land-use patterns within 1~km\textsuperscript{2} tiles. Second, the temporal modeling remains relatively coarse. Discrete month embeddings capture recurrent month-level correlations but fail to represent short-term variations driven by human activity, meteorological factors, heating and cooling demand, traffic intensity, or electricity use. Incorporating higher-resolution, continuous, or externally informed temporal representations may enable more fine-grained and dynamic emission prediction when such covariates are consistently available across cities. Finally, although inference relies solely on satellite imagery, pretraining still depends on street-view data and generated-text quality, restricting training in cities with limited multimodal resources. Reducing this dependency through self-supervised learning, synthetic street-level semantics, or cross-city knowledge transfer is an important direction for future research. Broader validation across more climatic zones, urban morphologies, and emission inventories is also needed before applying CarbonCLIP beyond the evaluated study cities.

\bibliographystyle{unsrt}
\bibliography{main}

\end{document}